\documentclass{article}

\usepackage{amsmath,amsfonts,bm}

\def\eqref#1{equation~\ref{#1}}

\def\1{\bm{1}}

\DeclareMathAlphabet{\mathsfit}{\encodingdefault}{\sfdefault}{m}{sl}
\SetMathAlphabet{\mathsfit}{bold}{\encodingdefault}{\sfdefault}{bx}{n}

\DeclareMathOperator*{\argmin}{arg\,min}

\usepackage{graphicx}
\usepackage{hyperref}
\usepackage{url}
\usepackage{algorithm}
\usepackage{algorithmic}
\renewcommand{\algorithmiccomment}[1]{#1}
\usepackage{xcolor,amsthm}

\usepackage{epigraph,booktabs}
\colorlet{shadecolor}{gray!90}

\usepackage{hyperref}

\usepackage[accepted]{icml2023}

\usepackage{amsmath}
\usepackage{amssymb}
\usepackage{mathtools}
\usepackage{amsthm}

\usepackage[capitalize,noabbrev]{cleveref}
\usepackage{mathtools}

\theoremstyle{plain}
\makeatletter
\newcommand{\@giventhat}[2]{\left(#1\;\middle|\;#2\right)}
\usepackage[textsize=tiny]{todonotes}

\icmltitlerunning{OCD: Learning to Overfit with Conditional Diffusion Models}

\begin{document}

\twocolumn[
\icmltitle{OCD: Learning to Overfit with Conditional Diffusion Models}



\icmlsetsymbol{equal}{*}

\begin{icmlauthorlist}
\icmlauthor{Shahar Lutati}{comp}
\icmlauthor{Lior Wolf}{comp}
\end{icmlauthorlist}
\icmlaffiliation{comp}{Blavatnik School of Computer Science, Tel Aviv University}

\icmlcorrespondingauthor{Shahar Lutati}{shahar761@gmail.com}
\icmlcorrespondingauthor{Lior Wolf}{wolf@cs.tau.ac.il}

\icmlkeywords{Machine Learning, ICML}

\vskip 0.3in
]



\printAffiliationsAndNotice{} 

\begin{abstract}

We present a dynamic model in which the weights are conditioned on an input sample $x$ and are learned to match those that would be obtained by finetuning a base model on $x$ and its label $y$. This mapping between an input sample and network weights is approximated by a denoising diffusion model. The diffusion model we employ focuses on modifying a single layer of the base model and is conditioned on the input, activations, and output of this layer. Since the diffusion model is stochastic in nature, multiple initializations generate different networks, forming an ensemble, which leads to further improvements. Our experiments demonstrate the wide applicability of the method for image classification, 3D reconstruction, tabular data, speech separation, and natural language processing. 
Our code is attached as supplementary material.%
\end{abstract}
\section{Introduction}

\epigraph{\normalsize Here is a simple local algorithm: For each testing pattern, (1) select the few training examples located in the vicinity of the testing pattern, (2) train a neural network with only these few examples, and (3) apply the resulting network to the testing pattern.}{\normalsize\citet{6797141}}
 
Thirty years after the local learning method in the epigraph was introduced, it can be modernized in a few ways. First, instead of training a neural network from scratch on a handful of samples, the method can finetune, with the same samples, a base model that is pre-trained on the entire training set. The empirical success of transfer learning methods~\citep{finetune2} suggests that this would lead to an improvement. 

Second, instead of retraining a neural network each time, we can learn to predict the weights of the locally-trained neural network for each set of input samples. This idea utilizes a dynamic, input-dependent architecture, also known as a hypernetwork~\citep{ha2016hypernetworks}.

Third, we can take the approach to an extreme and consider local regions that contain a single sample. During training, we finetune the base model for each training sample separately. In this process, which we call ``overfitting'', we train on each specific sample $s=(x,y)$ from the training set, starting with the weights of the base model and obtaining a model $f_{\theta_s}$. We then learn a model $g$ that maps between $x$ (without the label) and the shift in the weights of $f_{\theta_s}$ from those of the base model. Given a test sample $x$, we apply the learned mapping $g$ to it, obtain model weights, and apply the resulting model to $x$.

The overfitted models are expected to be similar to the base model, since the samples we overfit are part of the training set of the base model. 
As a result, it is likely that a diffusion process 
would be able to generate the weights of the fine-tuned networks. Recently, diffusion models, such as DDPM \citep{ho2020denoising} and DDIM \citep{ddim} were shown to be highly successful in generating perceptual samples \citep{diffusion_beats_gan,kong2021diffwave}. Here, we employ such models as hypernetworks, i.e., as means for conditionally generating network weights. 

In order to make the diffusion models suitable for predicting network weights, we make three adjustments. First, we automatically select a specific layer of the neural model and modify only this layer. This considerably reduces the size of the generated data and, in our experience, is sufficient for supporting the overfitting effect. Second, we condition the diffusion process on the input of the selected layer, its activations, and its output. Third, since the diffusion process assumes unit variance scale~\citep{ho2020denoising}, we learn the scale of the weight modification separately.

Similarly to other diffusion processes, our hypernetwork is initialized with normal noise and different initializations lead to slightly different results. Using this feature of the diffusion model, we  generate multiple models from the same instance and use the resulting ensemble technique to further improve the prediction accuracy. Our method is widely applicable, and  we evaluate it across five very different domains: image classification, image synthesis, regression in tabular data, speech separation, and few-shot NLP. In all cases, the results obtained by our method improve upon the baseline model to which our method is applied. Whenever the baseline model is close to the state of the art, the leap in  performance sets new state-of-the-art results.

\section{Related Work}

{\bf Local learning} approaches perform inference with models that are focused on training samples in the vicinity of each test sample. This way, the predictions are based on what are believed to be the most relevant data points. K-nearest neighbors, for example, is a local learning method. \citet{6797141} have presented a simple algorithm for adjusting the capacity of the learned model locally, and discuss the advantages of such models for learning with uneven data distributions. \citet{alpaydin1996local} combine multiple local perceptrons in either a cooperative or a discriminative manner, and \citet{zhang2006svm} combine multiple local support vector machines. These and other similar contributions rely on local neighborhoods containing multiple samples. The one-shot similarity kernel of \citet{wolf2009one} contrasts a single test sample with many training samples, but it does not finetune a model based on a single sample, as we do.


More recently, \citet{wang2021tent} employ local learning to perform single-sample domain adaptation (including robustness to corruption). The adaptation is performed through an optimization process that maximizes the entropy of the prediction provided for each test sample. Our method does not require any test-time optimization and focuses (on the training samples) on improving the accuracy of the ground truth label rather than label-agnostic confidence. 

\citet{alet2021tailoring} propose a method called Tailoring that employs, like our method, meta-learning to local learning. The approach is based on applying unsupervised learning on a dataset that is created by augmenting the test sample, in a way that is related to the adaptive instance normalization of \citet{huang2017arbitrary}. Our method does not employ any such augmentation and is based on supervised finetuning on a single sample.

Tailoring was tested on synthetic datasets with very specific structures, in a very specific unsupervised setting of CIFAR-10. Additionally, it was tested as a defense against adversarial samples, with results that fell short of the state of the art in this field. Since the empirical success obtained by Tailoring so far is limited and since there is no published code, it is not used as a baseline in our experiments.

As far as we can ascertain, all existing local learning contributions are very different from our work. No other contribution overfits samples of the training set, trains a hypernetwork for local learning, nor builds a hypernetwork based on diffusion models. 

{\bf Hypernetworks~\citep{ha2016hypernetworks}} are neural models that generate the weights of a second {\em primary} network, which performs the actual prediction task. Since the inferred weights are multiplied by the activations of the primary network, hypernetworks are a form of multiplicative interactions~\citep{jayakumar2020multiplicative}, and extend layer-specific dynamic networks, which have been used to adapt neural models to the properties of the input sample~\citep{klein2015dynamic,riegler2015conditioned}.  

Hypernetworks benefit from the knowledge-sharing ability of the weight-generating network and are therefore suited for meta-learning tasks, including few-shot learning~\citep{bertinetto2016learning}, continual learning~\citep{Oswald2020Continual}, and model personalization~\citep{shamsian2021personalized}. When there is a need to repeatedly train similar networks, predicting the weights can be more efficient than backpropagation.  Hypernetworks have, therefore, been used for neural architecture search~\citep{brock2018smash,zhang2018graph}, and hyperparameter selection~\citep{lorraine2018stochastic}. 

MEND by~\citet{mitchell2021fast} explores the problem of model editing for large language models, in which the model's parameters are updated after training to incorporate new data. In our work, the goal is to predict the label of the new sample and not to update the model. Unlike MEND, our method does not employ the label of the new sample.

{\bf Diffusion models\quad}
Many of the recent generative models for images \citep{ho2022cascaded,chen2020wavegrad, dhariwal2021diffusion} and speech \citep{kong2021diffwave,chen2020wavegrad} are based on a degenerate form of the Focker-Planck equation. \citet{sohl2015deep} showed that complicated distributions could be learned using a simple diffusion process. The Denoising Diffusion Probabilistic Model (DDPM) of \citet{ho2020denoising} extends the framework and presents high-quality image synthesis. 
\citet{ddim} sped up the inference time by an order of magnitude using implicit sampling with their DDIM method. \citet{watson2021learning} propose a dynamic programming algorithm to find an efficient denoising schedule and \citet{noise_scaling_nachamani} apply a learned scaling adjustment to noise scheduling. \citet{luhman2021knowledge} combined knowledge distillation with DDPMs.


The iterative nature of the denoising generation scheme creates an opportunity to steer the process, by considering the gradients of additional loss terms. The Iterative Latent Variable Refinement (ILVR) method~\citet{choi2021ilvr} does so for images by directing the generated image toward a low-resolution template. A similar technique was subsequently employed for voice modification~\citet{levkovitch2022zero}. Direct conditioning is also possible: \citet{imagen} generate photo-realistic text-to-image scenes by conditioning a diffusion model on text embedding;  \citet{amit2021segdiff} repeatedly condition on the input image to obtain image segmentation. In voice generation, the mel-spectrogram can be used as additional input to the denoising network~\citet{chen2020wavegrad,kong2021diffwave,liu2021diffsinger}, as can the input text for a text-to-speech diffusion model \citet{popov2021grad}. The conditioning we employ is of the direct type.

\section{Method}
Our method is based on a modified diffusion process. 
Denote the training dataset as $S=\{(x_i,y_i)\}_{i=1}^n$,
where $x_i$ are the data points in the dataset $S$, and $y_i$ are the associated labels. First, a base model, $f_\theta (x) = f(x,\theta)$ is trained over the entire dataset, $S$, where $\theta$ are the learned weights of the model when trained over the entire dataset. 

Next, for every training sample $s\in S$ we run fine-tuning based on that single sample to obtain the overfitted parameters (function) as $\theta_s$ ($f_{\theta_s}$). 
\begin{equation}
    \theta_s = \theta + \argmin_{\Delta}(\mathcal{L}(f(x_s,\theta + \Delta),y_s))\,,
\end{equation}
where $\mathcal{L}$ is the loss function that is minimized in the training of the base model, $x_s,y_s$ are the data point and label of sample $s$, and $\Delta$ is the weight difference obtained when finetuning the model. Finetuning is performed with three gradient descent iterations, and,  as shown in our runtime analysis, is typically much less computationally demanding than the training of the base network.

The meta-learning problem we consider is the one of learning a model $g$, which maps $x$ (the input domain of sample $s$), and potentially multiple latent representations of $x$ in the context of $f_\theta$, collectively denoted as $I(x)$, to a vector of weight differences, such that
\begin{equation}
    g(x,I(x)) = \theta_s - \theta
\end{equation}
 where $\theta$ are the base model's parameters trained over $S$, and $g(x,I(x))$ is a mapping function that maps the input, i.e., the $x$ part of $s$, and multiple latent representations of it, $I(x)$, to the desired shift in the model parameters.

{\bf Layer selection\quad}  Current deep neural networks can have millions or even billions of parameters. Thus, learning to modify all network parameters can be a prohibitive task. Therefore, we opt to modify, via function $g$, a single layer of $f_\theta$. 

To select this layer, we follow \citet{lutati2021hyperhypernetwork} and choose the layer that presents the maximal entropy of the loss, when fixing the samples $(x,y)\in s$, and perturbing the layer's parameters.

The layer that yields the highest entropy score when perturbed is a natural candidate for a fine-tuning algorithm, since a large entropy reflects that, near the layer's base parameters, the surface imposed by the loss function has a large variance. Thus, a small perturbation could result in a dramatic change in loss when the perturbation is in the right direction.

Denote the perturbed weights, in which only layer $L$ is perturbed, as $\theta^L$. The score used for selection is
\begin{equation}
Score=\frac{1}{|S|}\sum_{(x,y)\in S}\text{Entropy}_{\theta_L}(\mathcal{L}(f(x,\theta^L),y))\,,
\end{equation}
where $\mathcal{L}$ is the loss objective on which the function of $f_\theta$ is trained, and the entropy is computed over multiple draws of $\theta^L$. Since sampling does not involve a backpropagation computation, the process is not computationally demanding, and $10,000$ samples per each training sample $s=(x,y)$ are used.

The entropy, per each sample $s$, is computed by fitting a Gaussian Kernel Density Estimation (GKDE)~\citep{Silverman86} to the obtained empirical distribution of the loss function. 
The layer that obtains the highest mean entropy is selected.

{\bf The conditioning signal\quad} The latent representation, $I(x)$, has three components. Given a selected layer, $L$, we denote the input to this layer, when passing a sample $x$ to $f(x,\theta)$, as $i_L(x)$ and the activation of this layer as $a_L(x)$. We also use the output of the base function $f_\theta(x)$. $I(x)$ is given by the tuple
\begin{equation}
    I(x) = [i_L(x), a_L(x), f_\theta(x)]
    \label{eq:Ix_def}
\end{equation}

\subsection{Diffusion Process}

The goal of the diffusion process is to reconstruct $\Delta$, the difference between the fine-tuned weights $\theta_s$, and the base model weights, $\theta$.
The process iteratively starts a random $\Omega_T$, with the same dimensions as $\theta_s$. 
\begin{equation}
    \Omega_T \sim \mathcal{N}(0,1)
\end{equation}
Next, it iterates with $\Omega_t$, where $t$ is decreasing and is the diffusion step, and returns $\Omega_0$. After appropriate diffusion steps, $\Omega_0$ should be as close as possible to $\Delta$ for a given point $s$.

\begin{figure}
    \centering
    \includegraphics[width=0.9980\linewidth]{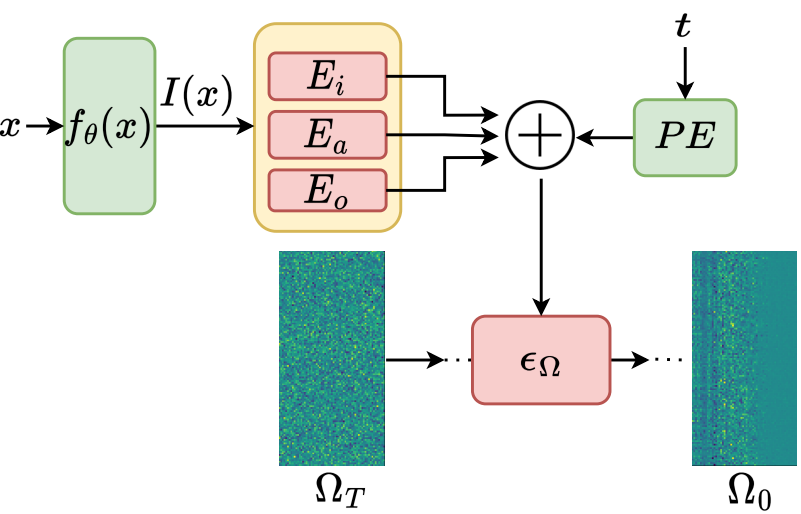}
    \caption{The diffusion process.     $x$ is the input of the base network, $f_\theta(x)$. $I(x)$ is a tuple of latent representations of $x$. 
    $E_i$,$E_a$, and $E_o$ are the input, activation, and output encoders, respectively, of the selected layer that is being modified. $t$ is the diffusion step, and $\Omega_t$ is the current diffusion estimation.}
    \label{fig:array_arc}
\end{figure}

The diffusion error estimation network, $\epsilon_\Omega$ is a function of the current estimation, $\Omega_t$, the latent representation tuple, $I(x)$, and the diffusion timestep, $t$. The last is encoded through a positional encoding network \citep{attention_is_all_u_need}, $PE$. All inputs, except for $\Omega_t$,  are combined into one vector: $e = PE(t) + E_{i}(i_L) + E_{a}(a_L) + E_{o}(f_\theta(x))$, where $E_i$, $E_{a}$, $E_{o}$ are the encodings of the layer input, layer activations, and network output. Note that most of the components of $e$ do not change during the diffusion process, and can be computed only once. This way, the conditioning overhead is reduced to a minimum. The conditional diffusion process is depicted in Fig~\ref{fig:array_arc}.

{\bf Training Phase\quad} The complete training procedure of $\epsilon_\Omega$ is depicted in Alg.~\ref{alg:diffalg}. The first phase is overfitting, using vanilla gradient descent over a single input-output pair, see line~1-5. The overfitting phase is not demanding, since the backpropagation is conducted only over the selected layer and a single sample. 

As stated in Sec.~\ref{subsec:Scale}, while regular diffusion assumes that the input has unit variance, when estimating network weights, scaling has a crucial impact. The normalization in line 6 ensures that the diffusion is trained over unit-variant input. We denote by $\Delta_{s}^\text{norm}$ the normalized difference between $\theta_s$ and the parameters $\theta$ of the base model.

Following \citet{ddim}, linear scheduling is used for the diffusion process, and $\beta_t$, $\alpha_t$, $\bar{\alpha}_t$, $\Tilde{\beta}_t$ are set in line~8. A training example is then sampled:
\begin{equation}
    \Omega_t = \sqrt{\bar{\alpha_t}}\Delta_{s}^{\text{norm}} + \sqrt{1-\bar{\alpha_t}}\epsilon\,,
\end{equation} where $\epsilon \sim \mathcal{N}(0,1)$ is normal noise. Since our goal is to recover the noiseless $\Delta_s^\text{norm}$, the objective is
\begin{equation}
    ||\epsilon - \epsilon_{\Omega}(\Omega_t,(I(x),t))||\,,
\end{equation} where $\epsilon_\Omega$ is the diffusion error estimating model defined above. A gradient step is taken in order to optimize this objective, see line.~10.

{\bf Inference Phase\quad} Given an unseen input $x$, $I(x)$ is computed using the base network $f(x,\theta)$ and is used for all calls to the diffusion network $\epsilon_\Omega$. The diffusion steps are listed in Appendix \ref{app:alg2}.


\begin{algorithm}[t]
    \caption{Training Algorithm. \\ \textbf{Input:} $S$ training set, $\theta$ base network parameters, $\mathcal{L}$ the loss of the primary task, $T$  diffusion steps\\
\textbf{Output:} $\epsilon_\Omega$ diffusion network (incl. $E_i$,$E_a$, $E_o$).}
\label{alg:diffalg}
\begin{algorithmic}[1]
\REPEAT
\STATE \text{sample $(x,y) \sim S$} and set $\theta_s = \theta$
\REPEAT
    \STATE Grad step on $\nabla \mathcal{L}(y,f_{\theta_s}(x))$ to update $\theta_s$  \label{lst:line:overfitting_alg_part}
\UNTIL{$\mathcal{L}(y,f_{\theta_s}(x))$ converges}\label{lst:loopends}
\STATE $\Delta_{s}^\text{norm} = \frac{\theta_{s}-\theta}{||\theta_s-\theta||}$ \label{lst:line:normalizing}
\STATE $t \sim Uniform({1...T})$, 
$\epsilon = N(\mathbf{0},\mathbf{1})$
\STATE $\beta_t = \frac{10^{-4}(T-t) + 10^{-2}(t-1)}{T-1}$,$\alpha_t = 1-\beta_t$,
$\bar{\alpha_t} = \Pi_{k=0}^{k=t} \alpha_k $ \label{scheduling}
\STATE $\Omega_t = \sqrt{\bar{\alpha_t}}\Delta_{s}^\text{norm} + \sqrt{1-\bar{\alpha_t}}\epsilon$  \label{sampling}
\STATE \text{Grad step on} $\nabla||\epsilon - \epsilon_\Omega(\Omega_t,(I(x),t))||$, updating $\epsilon_\Omega$, $PE$ and the components of $I$\label{lst:line:objective}
\UNTIL{$||\epsilon - \epsilon_\Omega(\Omega_t,(I(x),t))||$ converges}
\end{algorithmic}
\end{algorithm}


\subsection{Scale Estimation}
\label{subsec:Scale}

The Evidence Lower Bound (ELBO) used in \citet{ho2020denoising} assumes that the generated data has unit variance. In our case, where the generated data reflects a difference in the layer's weights, the scale of data presents considerable variation. Naturally, shifting the weights of a network by some vector $d$ or by some scale times $d$ can create a significant difference. 

We, therefore, use an MLP network $\rho(x,I(x))$ to estimate the appropriate scale factor, based on the same conditioning signal that is used for the network $\epsilon_\Omega$ that implements $g$ as a diffusion process. 
When learning network $\rho$, the following objective function is used
\begin{equation}
    \mathcal{L}_{\text{scale}} = \sum_{s=(x,y)\in S} 10\cdot log_{10}(\frac{|\rho(x,I(x))-\rho_s|^2}{\rho_s^2})\,,
\end{equation}
where $\rho_s = ||\theta_s-\theta||$.
The use of the log scale is due to the large variance of $\Delta$. When a data point $s$ is misclassified by the base model $f\theta$, the change in the model's weights, which results from finetuning, will be of a larger magnitude. In contrast, in the case of a sample  $s$ for which the output of $f_\theta$ matches the label, finetuning would result in a minor update.

\subsection{Architecture}
\label{sec:architecture}
Following  \citet{ho2020denoising}, the network $\epsilon_{\Omega}$ is a U-Net~\citep{unet}. Each resolution level has residual blocks and an attention layer. The bottleneck contains two attention layers. 
The weights $\theta$ are arranged as a matrix, with zeros added at the edges to create a square input matrix. The output of $\epsilon_{\Omega}$ is cropped to return the original dimensions of $\theta$.

The positional encoder is composed of stacked sine and cosine encodings, following \citet{attention_is_all_u_need}. The encoders of $i_L,a_L$ are both single fully-connected layers, with dimensions to match the positional embedding.  The encoder of the base network's output $f_\theta(x)$ depends on the output type. In the case of a classification network, where the output is a vector in $\mathbb{R^C}$, where $C$ is the number of classes, the encoder $E_O$ is a single fully-connected layer. This is also the case for our NLP experiments, which are multiclass classification experiments. In the case of image generation, the output image is first encoded using a VGG11 encoder \citep{vgg}, and then the latent representation is passed through a single fully-connected layer, again matching the dimension of the positional encoder. For speech separation, the estimated speech is first transformed to a spectogram with 1024 bins of FFT, and then encoded using the same VGG11.

\section{Experiments}

 In all experiments, the UNet $\epsilon_\Omega$ has 128 channels and five downsampling layers. The positional encoding, $PE$ has dimensions of 128x128. The Adam optimizer \citep{kingma2014adam} is used, with a learning rate of $10^{-4}$. A linear noise schedule is used based on \citet{ddim}, and the number of diffusion steps is 10. All experiments are repeated three times to report the standard deviation (SD) of the success metrics. 
 
In addition to the full method, we also show results for the network that overfits the test data, which serves as an upper bound that cannot be materialized without violating the train/test protocol. On some datasets we check to what extent selecting a single layer limits our results, by performing the overfitting process on all of the model weights. On all datasets, we ablate the scale component of our ``Overfit with Conditional Diffusion models'' (OCD) method, by estimating a fixed global scale factor $\bar \rho = \mathbb{E}_{s\in S}(\rho_s)$ as the mean value of the scale factor $\rho_s$ over the training set. An additional ablation selects the model $f_{\theta_s}$ of the training sample $s$ with the closest $x$ to the test sample. This ``nearest neighbor'' ablation can be seen as the simplest way to implement the concept of OCD. Finally, we present an ablation that selects the layer with the second-highest layer selection score, to evaluate the significance of the selection criterion.

As an additional baseline, we employ Tent's official implementation published by \citet{wang2021tent}. This baseline is employed in the vanilla classification experiments, which are similar in nature to the type of experiments Tent was previously evaluated for.

{\bf Image Classification\quad} Results for the MNIST dataset \citep{mnist}  are obtained with the LeNet5 architecture \citep{lent5}. The selected layer is the one next to the last fully connected layer, which, as can be seen in Appendix \ref{app:criterion} has the maximal entropy among LeNet5's layers. CIFAR10 images~\citep{cifar} are classified using GoogleNet~\citep{googlenet}. The selected layer was the last fully-connected layer, see Appendix \ref{app:criterion}. 
For both architectures, the three encoders $E_{L_i},E_{L_o},E_O$ are simple fully-connected layers, with dimensions to match the number of channels in the UNet (128). 

For classification experiments, we measure both the cross entropy (evaluated on the test set) and the test accuracy. As can be seen in Tab.~\ref{tab:results_image_classification_mnist}, our method reduces the CE loss by a factor of 8 in comparison to the base network and there is an improvement of 0.5\% in accuracy.  Ablating the scale prediction, the results considerably degrade in comparison to the full method. The Nearest-Neighbor ablation yields slightly better results than the base network. 

The ablation that selects an alternative layer results in performance that is similar to or slightly better than the base network. This is congruent with the small difference between fitting the selected layer and fitting all layers, which may suggest that much of the benefit of overfitting occurs in the selected layer.

On CIFAR10, our method improves classification accuracy from  92.9\% to 93.7\%. As in MNIST, much of the improvement is lost when the three ablations are run. 
In both MNIST and CIFAR, when using the ground truth to overfit a specific example, the accuracy becomes, as expected, 100\%. Considering the CE loss, overfitting the entire model instead of the selected layers yields only mild improvement (below the standard deviation for MNIST). This indicates that the added improvement gained by applying our method to all layers (and not just to the selected one) may not justify the additional resources required. 

Experiments were also conducted on the TinyImageNet dataset~\citep{le2015tiny}. The baseline network used is the (distilled) Data efficient transformer (DeiT-B/16-D) of \citet{touvron2022deit}. The selected layer is the weight of the head module. Our method is able to improve on the baseline network, achieving $90.8\%$ accuracy on the test set, and falls short of the current state of the art - which requires twice as many parameters - by less than 0.5\%. 

The weights in the OCD hypernetwork are obtained by a diffusion process, and multiple networks can be drawn for each sample $x$. When an ensemble of five classifiers is employed, using different initializations of noise in the diffusion process, the results surpass the current state of the art, yielding $92.00\%$ ($4.7\%$ better than the baseline model, and $0.65\%$ better than the current state of the art).

Tent is not competitive with OCD on any of the image classification datasets. In the results tables, in addition to the default of 10 Tent iterations, we provide results for 1 iteration, as provided by \citet{wang2021tent}, and for 30 iterations, for good measure. For all benchmarks and configurations, OCD is markedly better than Tent. While Tent improves over the baseline in both MNIST and TinyImageNet, in the latter case by almost a full percent, OCD, even without an ensemble, achieves more than 2.5 percent higher accuracy.

\begin{table*}[t]
\caption{Performance on classification tasks. CE=Cross Entropy}
\label{tab:results_image_classification_mnist}
\smallskip
\centering
 \begin{tabular}{@{}l@{~~}c@{~~}c@{~~}c@{~~}c@{}} 
 \toprule
  & \multicolumn{2}{c}{MNIST (LeNet5)} & \multicolumn{2}{c}{CIFAR10 (GoogleNet)}\\
 \cmidrule(lr){2-3}
  \cmidrule(lr){4-5}

Method &  Test-CE {$(\downarrow)$} & Accuracy \%{$(\uparrow)$} & Test-CE {$(\downarrow)$} & Accuracy \%{$(\uparrow)$} \\
\midrule
Base network &$0.080 \pm 0.009$ & $99.2 \pm 0.1$ &  $0.085 \pm 0.01$& $92.85 \pm 0.40$\\
{\color{shadecolor}Overfitting on test} &{\color{shadecolor} $0.002 \pm 0.0001$} &  {\color{shadecolor}$100$} & {\color{shadecolor}$0.075 \pm 0.005$} & {\color{shadecolor}$100$}\\
{\color{shadecolor}Overfitting on test (All Layers)}  &{\color{shadecolor}$0.002 \pm 0.0001$} & {\color{shadecolor} $100$} & {\color{shadecolor}$0.073 \pm 0.003$} &{\color{shadecolor} $100$}\\
OCD nearest neighbor ablation &$0.073 \pm 0.010$ & $99.3 \pm 0.1$ &  $0.082 \pm 0.02$& $93.03 \pm 0.40$\\
OCD no scaling ablation & $0.069 \pm 0.010$& $99.3 \pm 0.1$ & $0.084 \pm 0.02$ & $93.01\pm 0.35$\\
OCD alternative layer ablation & $0.078 \pm 0.010$ & $99.2 \pm 0.1$ & $ 0.084 \pm 0.01$ & $ 92.96\pm 0.27$\\
OCD (ours) & $\mathbf{0.010 \pm 0.006}$& $\mathbf{99.7 \pm 0.1}$ &  $\mathbf{0.080 \pm 0.01}$ & $\mathbf{93.68\pm 0.38}$\\
\midrule
Base network +  Tent (\citet{wang2021tent}, 1 iter) & $0.075\pm0.006$ & $99.4 \pm 0.1$ & $0.085 \pm 0.03$ & $92.8 \pm 0.40$\\
Base network +  Tent (\citet{wang2021tent}, 10 iters) & $0.073\pm0.004$ & 	$99.4 \pm 0.1$ & $0.083\pm0.04$ & $92.8 \pm 0.41$ \\
Base network +  Tent (\citet{wang2021tent}, 30 iters) & $0.073\pm0.004$ & 	$99.3 \pm 0.1$ & $0.083\pm0.03$ &$92.8 \pm 0.40$\\

 \bottomrule
\end{tabular}

\caption{Classification accuracy for TinyImageNet dataset.}
\label{tab:results_tinyImageNet}
\smallskip
\centering
 \begin{tabular}{lcc} 
 \toprule
 Method & \multicolumn{1}{c}{Test-CE {$(\downarrow)$}} & \multicolumn{1}{c}{Accuracy \%{$(\uparrow)$}}\\
    \midrule
Swin L/4 ~\citep{liu2021swin} & NA & $91.35$\\
DeiT-B/16-D~\citep{touvron2022deit} &$0.71\pm0.08$ & $87.29$ $\pm0.3$\\
DeiT-B/16-D + OCD & $0.65\pm 0.09$ & $90.80$ $\pm 0.35$\\
DeiT-B/16-D + OCD, ensemble of five & $0.65\pm0.09$  &$\mathbf{92.00}$ $\pm0.6$\\
\midrule
DeiT-B/16-D + Tent (\citet{wang2021tent}, 1 iter) & $0.71\pm0.05$ & $88.15 \pm 0.20$ \\
DeiT-B/16-D + Tent (\citet{wang2021tent}, 10 iters) & $0.70\pm0.02$  & $88.20 \pm 0.20$\\ 
DeiT-B/16-D + Tent (\citet{wang2021tent}, 30 iters) & $0.070\pm0.03$ &  $88.20 \pm 0.18$ \\
 \bottomrule
\end{tabular}
\end{table*}

 
\begin{table}[t]
\caption{MSE$\pm$SD, lower is better, for the TinyNeRF network.}
\label{tab:results_nerf}
\smallskip
\centering
 \begin{tabular}{@{}l@{~}c@{~}c@{~}c@{}} 
 \toprule
 Method & \multicolumn{1}{c}{Lego} & \multicolumn{1}{c}{Hotdog} &\multicolumn{1}{c}{Drums}\\
    \midrule
Base model &$.010 \pm .004$ &$.012 \pm .004$ &$.015 \pm .003$\\
{\color{shadecolor}Overfitting on test} & {\color{shadecolor} $.006 \pm .001$}& {\color{shadecolor}$.008 \pm .001$}& {\color{shadecolor}$.009\pm .003$}\\
OCD no scaling  & $.009 \pm .008$& $.012 \pm .005$& $.011 \pm .008$\\
OCD (ours) & $\mathbf{.008 \pm .006}$& $\mathbf{.009 \pm .004}$& $\mathbf{.010 \pm .004}$\\
 \bottomrule
\end{tabular}
\vspace{-.43cm}
\end{table}

\begin{figure*}[t]
\centering
  \begin{tabular}{@{}c@{~}c@{~}c@{}}
     \includegraphics[width=0.254\linewidth]{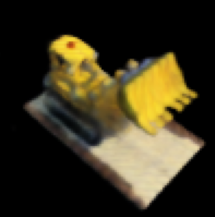} & 
          \includegraphics[width=0.254\linewidth]{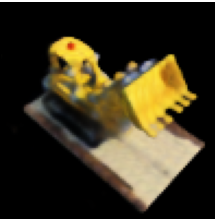} &
     \includegraphics[width=0.254\linewidth]{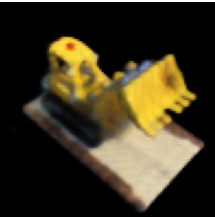}\\
     (a)&      (b)&     (c)\\
  \end{tabular}
    \caption{Sample TinyNeRF results (Lego model). More results can be found in Appendix~\ref{app:tinynerfresults}. 
    (a) Base model on a test view. (b) Same test view, overfitted using the ground truth  (c) OCD (ours).}
    \label{fig:array_sample}
    \vspace{-.3cm}
\end{figure*}

{\bf Image Synthesis\quad} We further tested our method on the image generation task of novel view synthesis, using a NeRF architecture \citep{nerf} and the ``Synthetic-Blender'' dataset. The Tiny-NeRF architecture \citep{tinynerf} employs an MLP network consisting of three fully-connected layers. The input is a 3D ray as a 5D coordinate (spatial location and viewing direction). The output is the corresponding emitted radiance. For each view, a batch of 4096 rays is computed, from which the interpolated image is synthesized. 

We experimented with three objects from the dataset: Lego, Hotdog, and Drums. For each object, a different TinyNeRF base model is trained over the corresponding training set. A single overfitting example is produced by considering a batch of 4096 rays from the same viewpoint.

Based on the data in Appendix \ref{app:criterion}, the first layer is selected. We, therefore, changed the layer-input encoder, $E_i$, such that the input image is first encoded by the VGG-11 encoder of \citet{vgg} (pretrained over ImageNet-1k), followed by a fully-connected layer, to match the dimensions of UNet channels. The encoders $E_a,E_o$ are simple fully-connected layers, with dimensions to match the number of channels in the UNet (128).

As can be seen in Tab.~\ref{tab:results_nerf}, our method improves the MSE by 31\% for the Lego model, by 25\% for Hotdog, and 16\% for Drums. Without input-dependent scaling, the performance is much closer to the base network than to that of our complete method. Sample test views are shown in Fig.~\ref{fig:array_sample} and in Appendix~\ref{app:tinynerfresults}. Evidently, our method improves both image sharpness and color palette, bringing the synthesized image closer to the one obtained by overfitting the test image.

{\bf Tabular Data\quad}
\citet{tabdata} have extensively benchmarked various architectures and tabular datasets. We use their simple MLP architecture as a base network (3 Layers). We were unable to reproduce the reported transformer, since the hyperparameters are not provided, and our resources did not allow us to run a neural architecture search, as \citet{tabdata} did. We run on two of the benchmarks listed: California Housing \citet{KELLEYPACE1997291} (CA), which is the first listed and has the least number of samples, and Microsoft LETOR4.0(MI) \citep{microsoft_dataset}, which is the last listed and has the largest number of samples. 

Based on the layer selection criterion, as depicted in Appendix \ref{app:criterion}, 
the first layer chosen for both datasets. As can be seen in Tab.~\ref{tab:results_tabular}, for CA the base MLP model falls behind ResNet. Applying our method, the simple architecture achieves better results. 

For MI when applying our method, the simple baseline achieves a record MSE of $0.743$, surpassing the current best record on this dataset, which is 0.745 \citep{Popov2020Neural}. 
The ablation that removes input-dependent scaling degrades the performance of the base network, emphasizing the importance of accurate scaling per sample.

\begin{table}[t]
    \centering
        \caption{Tabular benchmarks. $MSE\pm SD$, lower is better.}
    \label{tab:results_tabular}
    \smallskip
    \begin{tabular}{@{}l@{~}c@{~}c@{}}
         \toprule
         Method &\multicolumn{1}{c} {CA} & \multicolumn{1}{c} {MI}\\
        \midrule
        MLP  & $0.4990 \pm 0.0030$& $0.7470 \pm .0004$\\
        ResNet  & $0.4860 \pm  0.0030$&$0.7480 \pm .0003$\\
        {\color{shadecolor} Overfit MLP on test} & {\color{shadecolor}$0.4750 \pm  .0020$}& {\color{shadecolor}$0.7410 \pm .0003$}\\
        OCD + MLP no scale & $0.5000 \pm  .0030$& $0.7490 \pm  .0006$\\
        OCD + MLP (ours) & $\mathbf{0.4800 \pm  .0020}$&$\mathbf{0.7430 \pm  .0004}$\\
         \bottomrule
    \end{tabular}
\end{table}



\begin{table*}[t]
\begin{minipage}[c]{0.29\linewidth}
      \caption{Audio separation. DM: Dynamic Mixing}
    \label{tab:results_speech}
    \smallskip
       \centering
    \begin{tabular}{@{}l@{}c@{}} 
 \toprule
 Method & SI-SDRi[dB]($\uparrow$) \\%
    \midrule
SepIt \citep{sepit_paper} &$13.2 \pm 0.2$\\
\midrule
Baseline \citep{many_speak} &$12.7 \pm 0.1$\\
{\color{shadecolor} Overfit baseline on test} & {\color{shadecolor}$13.5 \pm 0.1$}\\
Baseline + OCD no scale & $12.8 \pm 0.3$\\
Baseline + OCD (ours) & $\mathbf{13.4 \pm 0.1}$\\
Baseline + OCD + DM (ours) & $\mathbf{13.9 \pm 0.1}$\\
 \bottomrule
\end{tabular}
 \end{minipage}%
\hfill
\begin{minipage}[c]{0.60\linewidth}
        \caption{Accuracy for few-shot NLP classification with 8 samples per class.}
    \label{tab:results_nlp}
    \smallskip
    \begin{tabular}{@{}l@{~}c@{~}c@{~}c@{~}c@{~}c@{~}c@{}} 
        \toprule
         Method & {SST-5} & {AmazonCF}&{CR}&{Emotion}&{Enron}&{Mean}\\
            \midrule
        T-few 3B~\citep{liu2022few} & $\mathbf{55.0}$&$19.0$&$92.1$&$\mathbf{57.4}$&$\mathbf{93.1}$&$63.3$\\
        \midrule
        SetFit~\citep{tunstall2022efficient} &$43.6$&$40.3$&$88.5$&$48.8$&$90.1$&$62.2$\\
        SetFit + OCD & $47.8$&$41.0$&$90.5$&$50.2$&$92.2$&${64.3}$\\
        Mean of 5 OCD instances& $47.9$&$41.0$&$90.3$&$50.2$&$92.3$&${64.3}$\\
        Ensemble of SetFit + OCD & $48.6$&$\mathbf{41.2}$&$\mathbf{91.2}$&$50.5$&$92.7$&$\mathbf{64.8}$\\        
         \bottomrule
        \end{tabular}
\end{minipage}
\end{table*}

{\bf Speech Separation\quad}
To tackle the task of single microphone speech separation of multiple speakers, \citet{VSUNS} introduce the Gated-LSTM architecture with MulCat block and \citet{many_speak} introduced a permutation-invariant loss based on the Hungarian matching algorithm, using the same architecture. \citet{sepit_paper}  further improved results for this architecture, by employing an iterative method based on a theoretical upper bound, achieving state-of-the-art results. 



The same backbone and Hungarian-method loss are used in our experiments, which run on the Libri5Mix dataset without augmentations, measuring the SI-SDRi score. The selected layer was the projection layer of the last MulCat block  (Appendix \ref{app:criterion}). The output of the Gated-LSTM is the separated sounds, and to encode it, we apply the audio encoding described in  Sec.~\ref{sec:architecture} to each output channel separately and concatenate before applying the linear projection to  $\mathbb{R}^{128}$.

As can be seen in Tab.~\ref{tab:results_speech}, applying our diffusion model over the Gated-LSTM model, we achieve $13.9dB$, surpassing current state-of-the-art results and approaching the results obtained by overfitting on the test data. The ablation that removes input-dependent scaling is much closer in performance to the base network than to our complete method.

{\bf Prompt-free Few-Shot NLP Classification\quad}
Recent few-shot methods have achieved impressive results in label-scarce settings. However, they are difficult to employ, since they require language models with billions of parameters and hand-crafted prompts. Very recently, \citet{tunstall2022efficient} introduced a lean hypernetwork architecture, named SetFit, for finetuning a pretrained sentence transformer over a small dataset (8 examples per class), and then employing logistic regression over the corresponding embedding. In our experiments, we apply OCD to the last linear layer of the SetFIT sentence transformer. The U-Net has 64 channels, with 5 downsample layers. The size of the last linear layer is $768x768$. $I(x) \in \mathbb{R}^{768x1}$ is the embedding of the Sentence Transformer.

As can be seen in Tab.~\ref{tab:results_nlp} using OCD and this lean architecture (with 110M parameters) outperforms, on average, the state-of-the-art model of \citet{liu2022few}, which has 3B parameters. The mean performance across the datasets is improved by 1.0\%  over the state-of-the-art and by 2.1\% over the SetFIT model we build upon. Recall that since the weights in the OCD hypernetworks are obtained by a diffusion process, one can draw multiple networks for each sample $x$. The variability that arises from the random initialization of the diffusion process allows us to use an ensemble. As can be seen in Tab.~\ref{tab:results_nlp}, when applying an ensemble of five inferences, the results of OCD further improve by 0.4\% (on average) to a gap of 1.5\% over the state-of-the-art. 

Interestingly, when testing the average network weights of the 5 networks, the accuracy is virtually identical to that obtained with a single OCD network. However, there is some reduction in Cross-Entropy for the mean network, see Appendix \ref{app:results_nlp_ce}).

{\bf Runtime\quad}
Tab.~\ref{tab:running_time} lists the measured runtime with and without OCD for both training and inference, on the low-end Nvidia RTX2060 GPU on which all experiments were run. Since overfitting each sample includes only 3 gradient descent steps over a single layer, most of the data collection results are shorter than the training time of the base model, with the exception of LeNet5 training that used a batch size of 16 versus batch size of 1 in the overfitting stage. Training the diffusion model is slower in most experiments than training the base model, except for the speech model, in which training the base model is lengthy. 

The inference overhead is almost constant across the six models, since it consists of running the UNet $\epsilon_\Omega$ ten times. The inference overhead due to a U-Net forward pass is 27[ms]  (there are ten such passes, one per diffusion step). In absolute terms,  the overhead incurred by OCD, even on a very modest GPU, is only a quarter of a second, while the results improve substantially. We note that the same $\epsilon_\Omega$ is used for all experiments, padding the actual weight matrix, which induces an unnecessary computational cost in the case of small networks.

\begin{table}[t]
    \caption{Runtime on a low-end Nvidia RTX2060GPU. Training includes the base model training, the collection of data for OCD by overfitting each sample in the training set, and the training of the diffusion model. The main overhead between the base model and the OCD one during inference is a result of running the UNet of the diffusion process for ten iterations.}
    \label{tab:running_time}
    \smallskip
    \centering
    \begin{tabular}{@{}l@{~}c@{~}c@{~}c@{~~}c@{~}c@{}}
    \toprule
    & \multicolumn{3}{c}{Training[hrs]}& \multicolumn{2}{c}{Inference[ms]}\\
    \cmidrule(lr){2-4}
    \cmidrule(lr){5-6}
         Architecture& Base & Overfit & Diffusion&Base & OCD\\
    \midrule
         LeNet5&0.5&1.0&6.0&15&288 \\
         GoogleNet&2.5&1.0&5.0&20&298 \\
         TinyNeRF&1.1&0.2&5.2&230&510 \\
         MLP&1.2&0.1&6.2&75&355 \\
         Gated-LSTM&30&2.5&6.8&450&730\\
         SetFit&0.2&0.1&5.1&125&401 \\
    \bottomrule
    \end{tabular}
\end{table}

\section{Conclusions}

We present what is, as far as we can ascertain, the first diffusion-based hypernetwork, and show that independently learning the scale is required for obtaining the best performance. The hypernetwork is studied in the specific context of local learning, in which a dynamic model is conditioned on a single input sample. 

The training samples for the hypernetwork are collected by finetuning a model on each specific sample from the training set used by the base model. By construction, this is only slightly more demanding than fitting the base model on the training set. More limiting is the size of the output of the hypernetwork, which for modern networks can be larger than the output dimensions of other diffusion processes. We, therefore, focus on a single layer, which is selected as the one most affected by weight perturbations. 

We tested our method extensively, tackling a very diverse set of tasks with the same set of hyperparameters. We are yet to find a single dataset or architecture on which our OCD method does not significantly improve the results of the baseline architecture.  

\section*{Acknowledgements}
The contribution of Shahar Lutati is part of a Ph.D. thesis
research conducted at Tel Aviv University.
This work was supported by a grant from the Tel Aviv University Center for AI and Data Science (TAD).

\bibliography{hypernet}
\bibliographystyle{icml2023}

\appendix
\newpage

\section{The inference time algorithm}
\label{app:alg2}
The steps of the inference algorithm are listed on Alg.~\ref{alg:inference}.

\begin{algorithm}[ht]
\caption{Inference Algorithm. \\ \textbf{Input:} $x$ input sample, $\theta$  the parameters of the base network, $\epsilon_\Omega$ diffusion network, T  diffusion steps.\\
\textbf{Output:} $g(x,I(x))$ estimated normalized ($\theta_s-\theta$) for $s$ associates with $x$.}
\label{alg:inference}
\begin{algorithmic}[1]
\STATE $t = T$
\STATE $\epsilon = N(\mathbf{0},\mathbf{1})$
\WHILE{$t \geq 0$} \label{lst:line:overfitting}
\STATE $\beta_t = \frac{10^{-4}(T-t) + 10^{-2}(t-1)}{T-1}$, $\alpha_t = 1-\beta_t$, $\bar{\alpha_t} = \Pi_{k=0}^{k=t} \alpha_k $, $\Tilde{\beta_t} = \frac{1-\bar{\alpha_{t-1}}}{1-\bar{\alpha_{t}}}\beta_t$
    \STATE $\Omega_{t-1} = \frac{\Omega_t - \frac{1-\alpha_t}{\sqrt{1-\bar{\alpha_t}}}\epsilon_\Omega(\Omega_t,I(x),t)}{\sqrt{\alpha_t}} + \mathbf{1}_{t>1}\sqrt{\Tilde{\beta_t}} $
\STATE $t = t-1$
\ENDWHILE\label{overfitting_while}
\STATE \textbf{return} $\Omega_0$
\end{algorithmic}
\end{algorithm}

\section{Layer selection plots}
\label{app:criterion}
\begin{figure*}[t]
  \begin{tabular}{ccc}
     \includegraphics[clip,width=0.32\linewidth,height=0.275\linewidth]{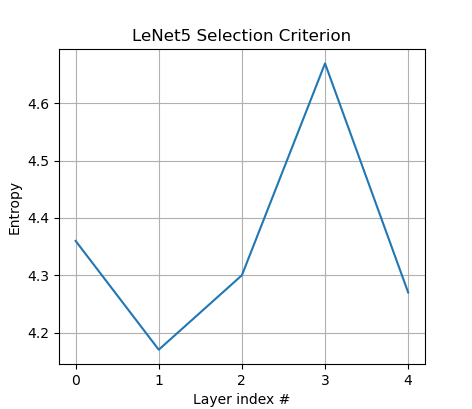} & 
          \includegraphics[clip,width=0.32\linewidth]{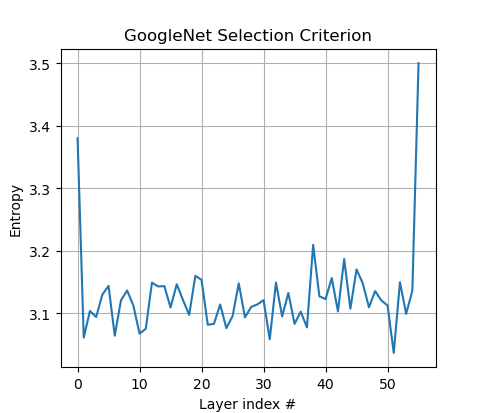} &
          \includegraphics[clip,width=0.32\linewidth,height=0.275\linewidth]{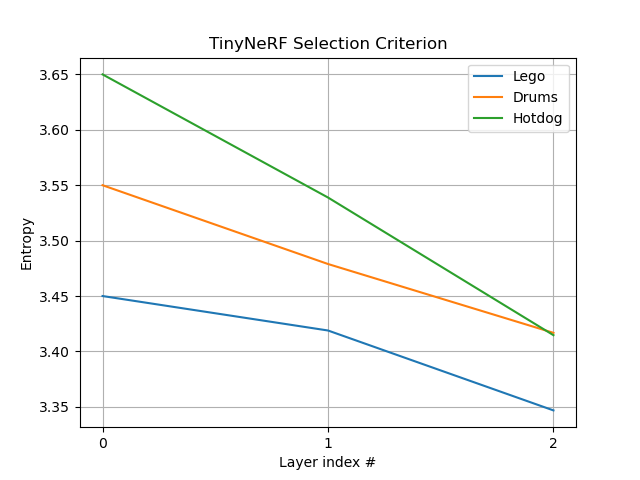}\\
          (a)&
     (b) &
     (c)\\
     \includegraphics[clip,width=0.32\linewidth,height=0.275\linewidth]{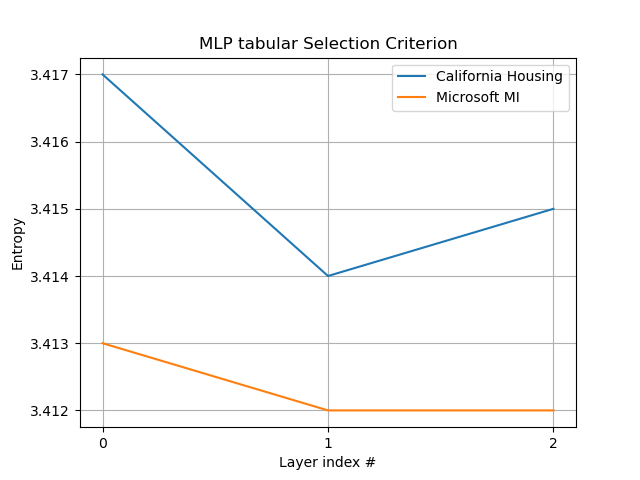} & 
     \includegraphics[clip,width=0.32\linewidth,height=0.275\linewidth]{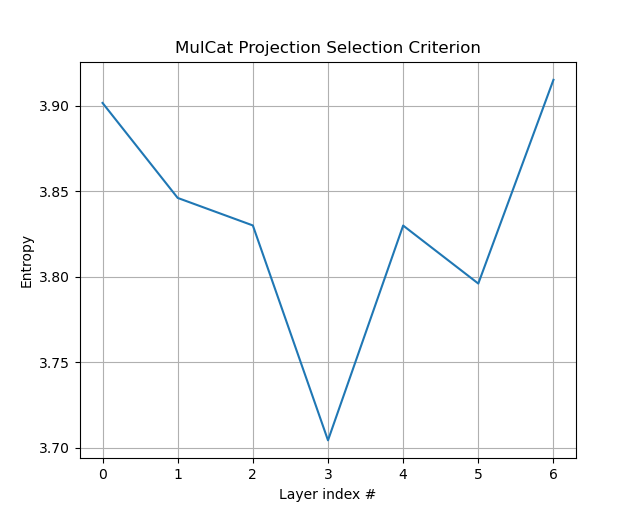} \\ 
     (d)&(e)
  \end{tabular}
    \caption{ Layer Selection Criterion for different experiments. (a) For LeNet5 on MNIST, the next to last Fully-Connected layer is selected since it has the maximal entropy. (b) For GoogleNet on CIFAR10, the last Fully-Connected layer is selected. (c) For TinyNeRF (three datasets), the first Fully-Connected layer is selected. 
    (d) For Tabular MLP the first layer is selected. (e) For MulCat the last projection layer is selected.}
    \label{fig:criterion}
\end{figure*}

Fig.~\ref{fig:criterion} depicts the layer selection criterion for various experiments.

\section{TinyNeRF sample results}

\label{app:tinynerfresults}
\begin{figure*}[t]
  \begin{tabular}{@{}ccc@{}}
     \includegraphics[width=0.32\linewidth]{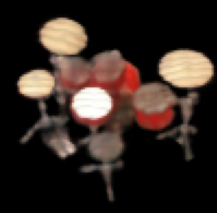} & 
          \includegraphics[width=0.32\linewidth]{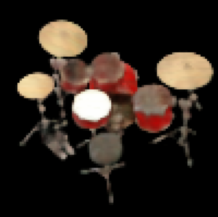} &
     \includegraphics[width=0.32\linewidth]{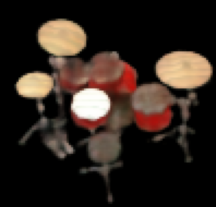}\\
     \includegraphics[width=0.32\linewidth]{lego_tf.png} & 
          \includegraphics[width=0.32\linewidth]{lego_tf_finetuned.png} &
     \includegraphics[width=0.32\linewidth]{lego_tf_diff.png}\\
          \includegraphics[width=0.32\linewidth]{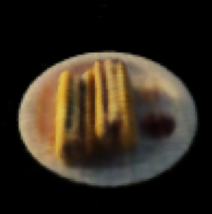} & 
          \includegraphics[width=0.32\linewidth]{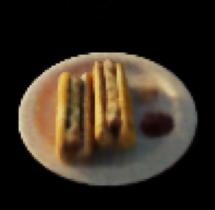} &
     \includegraphics[width=0.32\linewidth]{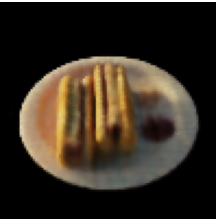}\\
     (a)&      (b)&     (c)\\
  \end{tabular}
    \caption{Sample TinyNeRF results on the Drums/Lego/Hotdog model. 
    (a) Base model on a test view. (b) Same test view, overfitted using the ground truth  (c) OCD (ours).}
    \label{fig:array_sample_1}
\end{figure*}

Fig.~\ref{fig:array_sample_1} presents sample results for the reconstruction of the Lego/Hotdog/Drums Model. The quality of the OCD results approaches that of the model that overfits on the test view.

\section{Cross entropy statistics for the NLP experiments}
\label{app:results_nlp_ce}
Table \ref{tab:results_nlp_ce} depicts the cross entropy loss for the NLP task of few-shot classification using SetFIT+OCD, when using one random initialization of the diffusion process or when averaging the networks obtained by five random initializations.

\begin{table*}[t]
    \centering
    \begin{tabular}{lccccc} 
         \toprule
          Method & \multicolumn{1}{c}{SST-5} & \multicolumn{1}{c}{AmazonCF}&\multicolumn{1}{c}{CR}&\multicolumn{1}{c}{Emotion}&\multicolumn{1}{c}{EnronSpam}\\
            \midrule
        One instance &$0.45\pm 0.19$&$0.65\pm 0.30$&$0.41\pm 0.11$&$0.63\pm 0.31$&$0.39\pm 0.20$\\
        Mean network of 5 instances & $0.42\pm 0.20$&$0.61\pm 0.25$&$0.39\pm 0.11$&$0.59\pm 0.32$&$0.38\pm 0.19$\\
         \bottomrule
        \end{tabular}
    \caption{The Cross-Entropy loss for the NLP task of few-shot classification with 8 samples per class using SetFIT + OCD, comparing one instance to the mean network obtained with 5 instances.}
    \label{tab:results_nlp_ce}
\end{table*}
\end{document}